%% file: main.tex
\documentclass[10pt,twocolumn,letterpaper]{article}

\usepackage[pagenumbers]{cvpr} 


\definecolor{cvprblue}{rgb}{0.21,0.49,0.74}
\usepackage[pagebackref,breaklinks,colorlinks,allcolors=blue]{hyperref}

\usepackage[vlined,linesnumbered,ruled]{algorithm2e}
\usepackage{multirow}
\usepackage{array}
\newcolumntype{C}[1]{>{\centering\arraybackslash}p{#1}}
\def\paperID{*****} 
\def\confName{CVPR}
\def\confYear{2025}

\title{2nd of the 5th PVUW MeViS-Audio Track: ASR-SaSaSa2VA}

\author{Zhiyu Wang$^{1}$\quad
        Xudong Kang$^{1}$\quad
        Shutao Li$^{1}$\quad  
        \\
    {$^{1}$Hunan University\quad}\\
}

\begin{document}
\maketitle
\input{sec/0_abstract}    
\input{sec/1_intro}

\input{sec/2_method}
\input{sec/3_exp}

{
    \small
    \bibliographystyle{ieeenat_fullname}
    \bibliography{main}
}


\end{document}

%% file: sec/0_abstract.tex
\begin{abstract}
Audio-based video object segmentation aims to locate and segment objects in videos conditioned on audio cues, requiring precise understanding of both appearance and motion. Recent audio-driven video segmentation methods extend MLLMs by fusing audio and visual features for end-to-end localization. Despite their promise, these approaches are computationally intensive, struggle with aligning temporal audio cues to dynamic video content, and depend on large paired audio-video datasets. To address these challenges, we present ASR-SaSaSa2VA, a resource-efficient framework for audio-guided video segmentation. The key idea is to convert audio inputs into textual motion descriptions via automatic speech recognition (ASR) models and then leverage pre-trained text-based referring video segmentation models (e.g., SaSaSa2VA) for pixel-level predictions. To further enhance robustness, we incorporate a no-target expression detection module, implemented by a fine-tuned audio-based MLLM, which filters out audio clips that do not refer to any target object. This design allows the system to exploit strong pre-trained models while effectively handling ambiguous or irrelevant audio inputs. Our approach achieves a final score of 80.7 in the 5th PVUW Challenge (MeViS-v2-Audio track), earning the second-place ranking.
\end{abstract}

%% file: sec/1_intro.tex
\section{Introduction}
\label{sec:intro}

\subsection{Challenge Description}

Pixel-level understanding of visual content is a core problem in computer vision~\cite{han2022survey}, encompassing tasks such as instance segmentation~\cite{guo2025audio} and semantic segmentation~\cite{wang2025vidseg}. While much progress has been made on static images, real-world applications are inherently dynamic, requiring methods that operate on video sequences~\cite{ding2023mose, ding2025mosev2}. Recognizing this, the Pixel-level Video Understanding in the Wild (PVUW) Challenge aims to advance research on video segmentation in unconstrained environments, where objects undergo complex motion, occlusions, and appearance variations over time. By providing large-scale, carefully curated datasets and standardized evaluation protocols, PVUW offers a benchmark for methods that can interpret dynamic scenes at a fine-grained, pixel-level resolution. The 5th PVUW Challenge~\cite{pvuw2026}, organized in conjunction with CVPR 2026, consists of three distinct tracks targeting different aspects of video understanding:

\begin{itemize}
    \item \textbf{Track 1: Complex Video Object Segmentation (MOSEv2).} This track focuses on segmenting and tracking objects in videos captured under challenging conditions, including cluttered backgrounds, occlusions, and small or reappearing objects. It tests the ability of algorithms to robustly maintain object identity across frames in realistic scenarios.

    \item \textbf{Track 2: Text-based Referring Motion Expression Video Segmentation (MeViS-v2-Text).} This track emphasizes motion-driven referring video segmentation guided by textual descriptions. Unlike conventional referring video datasets that rely heavily on static object attributes, MeViS-Text requires models to leverage dynamic motion cues both in the visual content and in the language expressions.

    \item \textbf{Track 3: Audio-based Referring Motion Expression Video Segmentation (MeViS-v2-Audio).} This novel track investigates the feasibility of locating and segmenting objects in video using audio cues that describe their motion. The task introduces the challenge of reasoning across modalities, as audio signals must be mapped to pixel-level visual predictions over time.
\end{itemize}

\subsection{MeViSv2-Audio Track}

In this work, we focus on the MeViS-Audio Track, which presents unique challenges compared to conventional text-driven referring video segmentation. Audio descriptions encode motion information in a temporal and often ambiguous manner, making direct localization of target objects non-trivial. 

Recently, Multi-modal Large Language Models (MLLMs) have demonstrated remarkable progress in image and video understanding~\cite{yin2024survey, rao2023DCD}, achieving holistic scene comprehension, object attribute recognition, action detection, and inter-object reasoning. For text-based referring video segmentation, models such as Sa2VA~\cite{yuan2025sa2va} and SaSaSa2VA~\cite{niu20251st} have established strong baselines. In the audio-based setting, recent approaches extend these ideas by integrating audio encoders with MLLMs~\cite{ding2025mevis, xu2025qwen3}, often training them jointly to enable the model to extract motion cues from audio signals and align them with corresponding visual features. Such methods typically encode audio using pre-trained speech or audio representation models and fuse them with video representations in a multi-modal architecture to achieve end-to-end localization and segmentation.

Despite these advances, these end-to-end audio-based video segmentation methods face several practical challenges. First, end-to-end training of audio-video models requires substantial computational resources, often limiting their applicability in resource-constrained settings. Second, aligning temporal audio cues with dynamic visual content remains difficult, particularly in long videos or when multiple objects produce overlapping sounds. Third, current audio-video methods often rely on large-scale paired audio-video datasets, which are scarce, leading to suboptimal generalization when models encounter unseen motions or background noise. These limitations motivate the exploration of resource-efficient pipelines that leverage pre-trained models while effectively incorporating audio-driven motion cues.

To address these challenges, we present \textbf{ASR-SaSaSa2VA}, a simple yet effective framework that leverages pre-trained text-based referring video segmentation models (e.g., SaSaSa2VA~\cite{niu20251st}) in conjunction with automatic speech recognition (ASR) models to convert audio inputs into textual motion descriptions. By transforming the audio modality into a textual representation, our approach enables the direct utilization of powerful pre-trained text-based referring video segmentation models, resulting in a practical and resource-efficient solution for audio-guided video object segmentation. Furthermore, to handle \emph{no-target expressions}, we fine-tune an audio-based MLLM (e.g., Qwen2.5-Omni) to determine whether a given audio clip refers to any valid target in the video. This auxiliary module improves the robustness of the overall system by filtering out ambiguous or irrelevant audio inputs before segmentation.

\begin{table}[t!]
    \centering
    \renewcommand{\arraystretch}{1.25}
    \begin{tabular}{r|l|ccc|c}
    \toprule
    Rank & Team & $\mathcal{J\&F}$ & N-acc & T-acc & Final \\
    \midrule
    \# 1 & yahooo  & 67.0 & 89.4 & 97.7 & 84.7 \\
    \rowcolor{lightgray} \# 2 & HNU-VPAI   & 63.9 & 83.3 & 94.9 & 80.7 \\
    \# 3 & csjihwanh  & 53.9 & 69.7 & 81.6 & 68.4 \\
    \# 4 & vvv666  & 47.2 & 12.1 & 97.7 & 52.3 \\
    \# 5 & liyiying  & 47.7 & 9.1 & 96.5 & 51.1 \\
    \bottomrule
    \end{tabular}
    \caption{\textbf{Leaderboard of the 5th PVUW Challenge (MeVisV2-Audio track) in CVPR 2026.} Our \textbf{HNU-VPAI} team achieves a final score of \textbf{80.7} and ranks second place.}
    \label{tab:leaderboard}
\end{table}

\begin{figure*}[t]
  \centering
  \includegraphics[width=1.0\linewidth]{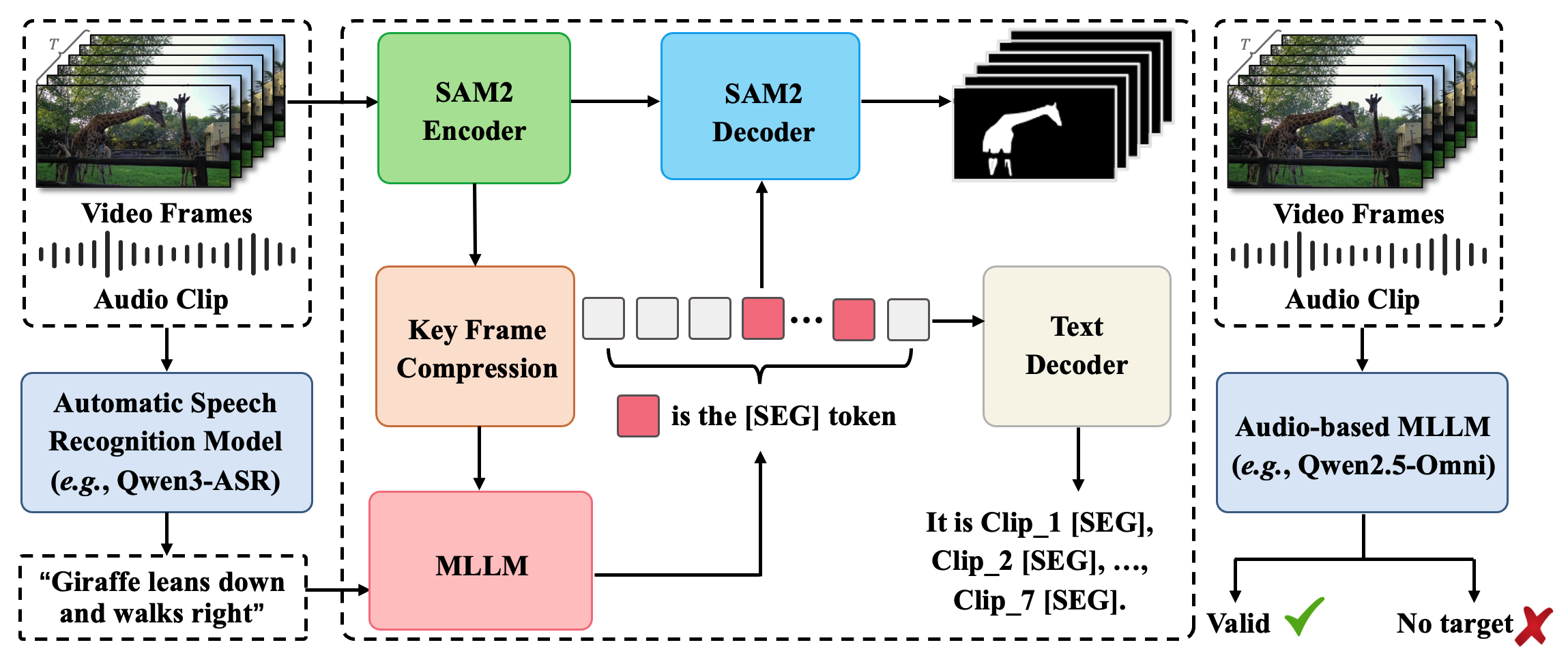}
  \caption{\textbf{Overview of our method.}  ASR-SaSaSa2VA decomposes the task into three stages: (1) converting audio inputs into textual motion descriptions via automatic speech recognition models, (2) performing text-based referring video segmentation using a pre-trained model, and (3) filtering out invalid queries through a no-target expression detection module.}
  \label{fig:sa}
\end{figure*}

With the proposed design, ASR-SaSaSa2VA demonstrates strong performance on the audio-guided video segmentation task. As shown in~\cref{tab:leaderboard}, our method achieves a final score of \textbf{80.7}, ranking second in the competition. This result highlights the effectiveness of transforming audio inputs into textual representations for leveraging powerful pre-trained video-language models, as well as the benefit of incorporating the no-target detection module for improved robustness.

%% file: sec/2_method.tex





\section{ASR-SaSaSa2VA}
\label{sec:method}

In this section, we present \textbf{ASR-SaSaSa2VA}, a modular framework for audio-guided video object segmentation. As illustrated in Fig.~\ref{fig:sa}, our method decomposes the task into three stages: (1) converting audio inputs into textual motion descriptions via automatic speech recognition (ASR) models, (2) performing text-based referring video segmentation using a pre-trained model, and (3) filtering out invalid queries through a no-target expression detection module. This design enables us to effectively leverage powerful pre-trained models while avoiding the high computational cost of end-to-end audio-video training.

\subsection{Automatic Speech Recognition}

Given an input video $\mathcal{V}$ and its associated audio signal $\mathcal{A}$, our first step is to convert the audio modality into a textual representation. Specifically, we employ an off-the-shelf automatic speech recognition (ASR) model to transcribe $\mathcal{A}$ into a text sequence $\mathcal{T}$:
\begin{equation}
    \mathcal{T} = \text{ASR}(\mathcal{A}).
\end{equation}

In this work, we adopt \textbf{Qwen3-ASR}~\cite{shi2026qwen3} as our ASR backbone. Qwen3-ASR is a family of all-in-one speech recognition models built upon the audio foundation model Qwen3-Omni, supporting multilingual ASR and language identification across diverse real-world scenarios. It is trained on large-scale speech corpora and demonstrates strong transcription quality and robustness, achieving competitive performance with state-of-the-art systems while maintaining high efficiency. Such properties make it particularly suitable for our pipeline, where reliable transcription of motion-related audio cues is critical.

The resulting transcription $\mathcal{T}$ serves as a proxy for motion expressions originally encoded in the audio signal. This transformation bridges the modality gap between audio and text, allowing us to directly utilize existing text-based referring video segmentation models.

\subsection{Text-based Video Segmentation}

Given the video $\mathcal{V}$ and the transcribed text $\mathcal{T}$, we adopt a pre-trained text-based referring video segmentation model to predict pixel-level masks of the target object. In this work, we build upon \textbf{SaSaSa2VA}~\cite{niu20251st}, which extends the Sa2VA framework~\cite{yuan2025sa2va} with enhanced temporal modeling capability.

\paragraph{Sa2VA Overview.}
Sa2VA consists of a multi-modal large language model (MLLM) and a segmentation backbone (SAM2~\cite{ravi2024sam}). The MLLM takes video frames and text instructions as input and generates textual responses. When segmentation is required, the model produces a special token \texttt{[SEG]}, whose hidden representation serves as an implicit prompt for segmentation. This prompt is then fed into SAM2 to generate object masks for the video.

Concretely, the MLLM (e.g., InternVL 2.5~\cite{chen2024expanding}) encodes video frames into visual tokens and fuses them with text tokens. Through autoregressive decoding, it outputs responses containing \texttt{[SEG]} tokens. The hidden state of each \texttt{[SEG]} token is projected and used as input to SAM2, which generates high-quality masks for selected frames and propagates them across the video sequence.

\paragraph{SaSaSa2VA Enhancement.}
SaSaSa2VA improves upon Sa2VA by addressing its limitations in temporal modeling. In particular, Sa2VA processes only a small number of sampled frames and relies on a single \texttt{[SEG]} token to represent the entire video, which restricts its ability to capture long-range temporal dependencies.

To overcome this, SaSaSa2VA introduces a \emph{Key Frame Compression (KFC)} strategy and multiple \texttt{[SEG]} tokens. Specifically, a video is divided into multiple clips, each containing a key frame and a compressed representation of neighboring frames. This design enables the MLLM to access more global temporal context while maintaining computational efficiency. Meanwhile, assigning one \texttt{[SEG]} token per clip allows the model to generate more fine-grained segmentation prompts, improving robustness to temporal variations in object motion and appearance.

\subsection{No-target Expression Detection}

In this challenge, not all audio inputs correspond to valid target objects in the video. To improve robustness, we introduce a no-target expression detection module that determines whether a given audio clip contains a valid referring expression.

Specifically, we adopt \textbf{Qwen2.5-Omni} as the backbone, an end-to-end multimodal large language model capable of jointly modeling text, images, audio, and video. Benefiting from its unified multimodal representation and strong cross-modal reasoning ability, Qwen2.5-Omni is naturally well-suited for understanding the semantic alignment between audio descriptions and visual content. 

To adapt the model to our task, we fine-tune Qwen2.5-Omni using a parameter-efficient LoRA strategy to perform binary classification:
\begin{equation}
    y = f_{\text{cls}}(\mathcal{A}, \mathcal{V}),
\end{equation}
where $y \in \{0,1\}$ indicates whether the audio describes a target object present in the video.

If $y = 0$, the input is classified as a no-target expression, and the segmentation stage is skipped. Otherwise, the transcribed text $\mathcal{T}$ is passed to the segmentation model. This lightweight adaptation enables effective filtering of ambiguous or irrelevant audio inputs, reducing false positives and improving the overall robustness of the system without introducing significant computational overhead.

%% file: sec/3_exp.tex
\section{Experiments}

\subsection{Implementation Details}

For text-based video segmentation, we adopt the SaSaSa2VA-4B and SaSaSa2VA-26B models. The Key Frame Compression (KFC) strategy used in SaSaSa2VA follows the Uniform+ configuration. The audio modality is processed using Qwen3-ASR-1.7B, an off-the-shelf automatic speech recognition model capable of transcribing audio from multiple languages with high accuracy. To handle no-target expressions, we employ Qwen2.5-Omni-7B fine-tuned with LoRA for binary classification.

\subsection{Main Results}

The performance of our ASR-SaSaSaVA framework on the 5th PVUW Challenge (MeViS-v2-Audio track) is summarized in Table~\ref{tab:leaderboard}. Our method achieves a final score of 80.7, securing second place in the competition. Specifically, it attains a $\mathcal{J\&F}$ of 63.9, a N-acc of 83.3, and a T-acc of 94.9, consistently outperforming most other participating teams.

These results highlight the effectiveness of our approach in several aspects. First, transforming audio inputs into textual motion descriptions allows us to leverage strong pre-trained text-based referring video segmentation models, achieving high-quality pixel-level predictions without the need for end-to-end audio-video training. Second, the integration of a no-target expression detection module ensures robust handling of ambiguous or irrelevant audio clips, reducing false positives and improving overall reliability.

\subsection{Ablation Study}
\label{sec:ablation_study}

\noindent\textbf{Impact of ASR Model.} We evaluate the effect of different ASR models for audio transcription while keeping other components fixed. Specifically, we use SaSaSa2VA-4B as the text-based video segmentation model and do not apply the no-target expression detection module in this experiment. We compare Qwen3-ASR-1.7B with a baseline Fun-ASR-0.8B model. As shown in Table~\ref{tab:ablation_asr}, using Qwen3-ASR provides improvements in $\mathcal{J\&F}$, indicating that higher-quality transcriptions yield more accurate textual motion cues for downstream segmentation. 

\begin{table}[t!]
    \centering
    \renewcommand{\arraystretch}{1.25}
    \begin{tabular}{l|ccc}
    \toprule
    ASR Model & $\mathcal{J\&F}$ & N-acc & T-acc \\
    \midrule
    Fun-ASR-0.8B & 45.4 & 8.4 & 95.8 \\
    Qwen3-ASR-1.7B & 46.6 & 7.6 & 96.1 \\
    \bottomrule
    \end{tabular}
    \caption{\textbf{Ablation study on the ASR model.} Here, the text-based video segmentation model is fixed to SaSaSa2VA-4B and no-target detection is not used. Using Qwen3-ASR improves most metrics compared to Fun-ASR.}
    \label{tab:ablation_asr}
\end{table}

\noindent\textbf{Impact of Text-based Video Segmentation Model.} Next, we evaluate the effect of model scale by comparing SaSaSa2VA-4B and SaSaSa2VA-26B. As reported in Table~\ref{tab:ablation_seg}, the larger model improves $\mathcal{J\&F}$ by several points, suggesting that increased model capacity enhances the alignment between textual motion cues and visual features.

\begin{table}[t!]
    \centering
    \renewcommand{\arraystretch}{1.25}
    \begin{tabular}{l|ccc}
    \toprule
    Segmentation Model & $\mathcal{J\&F}$ & N-acc & T-acc \\
    \midrule
    SaSaSa2VA-4B & 46.6 & 7.6 & 96.1 \\
    SaSaSa2VA-26B & 49.4 & 7.6 & 97.3 \\
    \bottomrule
    \end{tabular}
    \caption{\textbf{Ablation study on the text-based video segmentation model.} Larger model size improves performance across all metrics.}
    \label{tab:ablation_seg}
\end{table}

\noindent\textbf{Impact of No-target Expression Detection.} Finally, we analyze the contribution of the no-target expression detection module. As shown in Table~\ref{tab:ablation_notarget}, including this module significantly improves N-acc and reduces false positives, while also slightly improving $\mathcal{J\&F}$ by filtering irrelevant audio queries.

\begin{table}[t!]
    \centering
    \renewcommand{\arraystretch}{1.25}
    \begin{tabular}{l|ccc}
    \toprule
    No-target Detection & $\mathcal{J\&F}$ & N-acc & T-acc \\
    \midrule
    Without & 49.4 & 7.6 & 97.3 \\
    With    & 63.9 & 83.3 & 94.9 \\
    \bottomrule
    \end{tabular}
    \caption{\textbf{Ablation study on the no-target expression detection module.} The module improves robustness by filtering irrelevant audio inputs.}
    \label{tab:ablation_notarget}
\end{table}

\section{Conclusion}
\label{sec:conclusion}

In this report, we present ASR-SaSaSaVA, a framework for audio-driven referring video object segmentation that effectively leverages pre-trained text-based segmentation models and audio understanding models. By converting audio signals into textual motion descriptions via Qwen3-ASR and incorporating a no-target expression detection module based on Qwen2.5-Omni, our approach addresses the practical challenges of computational efficiency, cross-modal alignment, and robustness to irrelevant audio inputs. Our method demonstrates strong performance in the 5th PVUW Challenge (MeViS-v2-Audio track), achieving a final score of 80.7 and ranking second on the leaderboard.